\documentclass[letterpaper, 10 pt, conference]{ieeeconf}  

\IEEEoverridecommandlockouts                              

\usepackage{amsmath} 
\usepackage{amssymb}  
\usepackage{mathtools} 
\DeclarePairedDelimiter{\norm}{\lVert}{\rVert}
\DeclareMathOperator{\arctantwo}{arctan2}
\usepackage{color}
\usepackage{import}
\usepackage{siunitx}
\usepackage{subfig}
\usepackage{ifthen} 
\usepackage{booktabs}
\usepackage[bookmarks=true,hidelinks]{hyperref}

\title{\LARGE \bf
Towards Predicting Fine Finger Motions from Ultrasound Images via Kinematic Representation
}

\author{Dean Zadok$^{1}$, Oren Salzman$^{1}$, Alon Wolf$^{2}$ and Alex M. Bronstein$^{1}$
\thanks{$^{1}$Department of Computer Science, Technion, Haifa, Israel
        {\tt\small \{deanzadok,osalzman,bron\}@cs.technion.ac.il}}%
\thanks{$^{2}$Department of Mechanical Engineering, Technion, Haifa, Israel
        {\tt\small alonw@me.technion.ac.il}}%
}

\newcommand{\calS}{\ensuremath{\mathcal{S}}}
\newcommand{\calX}{\ensuremath{\mathcal{X}}}

\newcommand{\E}{\mathbb{E}}

\newcommand{\probP}{\text{I\kern-0.15em P}}

\newcommand{\Cpp}{C\raise.08ex\hbox{\tt ++}}

\newcommand{\SF}{\textsf{SF}}
\newcommand{\MF}{\textsf{MF}}
\newcommand{\CBMF}{\textsf{CBMF}}

\newcommand{\MSE}{\text{MSE}}
\newcommand{\BCE}{\text{BCE}}

\newcommand{\ignore}[1]{}

\newboolean{HIDENOTES}
\setboolean{HIDENOTES}{false}

\ifthenelse{\boolean{HIDENOTES}}{
    \newcommand{\OS}[1]{{}}
    \newcommand{\DZ}[1]{{}}
    \newcommand{\AB}[1]{{}}
    \newcommand{\AW}[1]{{}}
    \newcommand{\CONT}[1]{{}}
}{
    \newcommand{\OS}[1]{\textcolor{red}{#1}}
    \newcommand{\DZ}[1]{\textcolor{blue}{#1}}    
    \newcommand{\AB}[1]{{\textcolor{green}{#1}}}    
    \newcommand{\AW}[1]{{\textcolor{yellow}{#1}}}

}

\newcommand{\ARXIVappendixref}{the appendix~(Sec.~\ref{sec:app})}
\newcommand{\ICRAappendixref}{the extended version of this paper~\cite{DBLP:journals/corr/abs-2202-05204}}

\newboolean{isARXIVversion}

\begin{document}

\setboolean{isARXIVversion}{true}
\newcommand{\RefAppendix}{\ifthenelse{\boolean{isARXIVversion}}{\ARXIVappendixref}{\ICRAappendixref}}

\maketitle
\thispagestyle{empty}
\pagestyle{empty}


\begin{abstract}

A central challenge in building robotic prostheses is the creation of a sensor-based system able to read physiological signals from the lower limb and instruct a robotic hand to perform various tasks. Existing systems typically perform discrete gestures such as pointing or grasping, by employing electromyography (EMG) or ultrasound (US) technologies to analyze muscle states. 
While estimating finger gestures has been done in the past by detecting prominent gestures, we are interested in detection, or inference, done in the context of fine motions that evolve over time. Examples include motions occurring when performing fine and dexterous tasks such as keyboard typing or piano playing.
We consider this task as an important step towards higher adoption rates of robotic prostheses among arm amputees, as it has the potential to dramatically increase functionality in performing daily tasks. 
To this end, we present an end-to-end robotic system, which can successfully infer fine finger motions.
This is achieved by modeling the hand as a robotic manipulator and using it as an intermediate representation to encode muscles' dynamics from a sequence of US images.
We evaluated our method by collecting data from a group of subjects and demonstrating how it can be used to replay music played or text typed. To the best of our knowledge, this is the first study demonstrating these downstream tasks within an end-to-end system.

\end{abstract}


\section{INTRODUCTION}

Bionic hands or robotic prostheses were sought after for decades yet the first commercially-available robotic prostheses only became available in~2007~\cite{connolly2008prosthetic} and a fully-functioning prosthesis enabling all range of daily tasks is still out of reach.
Existing robotic prostheses are typically connected to the limb of an amputee (or to a subject with severe phalanx or palm deformations) when the subject's muscles still respond to electric potentials generated by the brain~\cite{herberts1969myoelectric}. 
These electric potentials allow inferring different finger motions by placing a non-invasive sensor on the residual limb, having subjects attempt these motions or gestures, and then discriminating between the different types of recorded electrical signals~\cite{DBLP:journals/corr/abs-1908-10522, said2019customizable}.
This sensing technology, called electromyography (EMG), allows the reproduction of grasp flexion and extension yet is unable to reproduce finer motions typically involved in daily tasks.

Recently, it was shown that replacing EMG with ultrasound (US) imaging that captures the muscles' morphological state allows for better differentiation between discrete gestures or to classify full-finger flexion~\cite{DBLP:journals/corr/abs-1808-06543}. 
These images are obtained by placing an US probe on the residual limb and its efficacy is based on the fact that the muscles generate different deformation patterns for different actions, and this behavior can be generalized among different subjects~\cite{DBLP:journals/corr/abs-1808-06543, yan2019lightweight, DBLP:journals/titb/YangZZHL19, yang2020simultaneous}.
Arguably, the most popular approach to infer motions from (high-dimensional) US images is via data-driven methods~\cite{yan2019lightweight, yang2020simultaneous, DBLP:conf/chi/McIntoshMFP17}. 
However, to the best of our knowledge, these methods still fall short of fully predicting dexterous hand motions or fine finger motions that a subject intended to perform. 
Predicting such motions, which is the focus of our work, is instrumental to the fine control of robotic hands and, consequently, to increase adoption rates of robotic prostheses.

\begin{figure*}[t]
\centering
\begingroup%
  \makeatletter%
  \providecommand\color[2][]{%
    \errmessage{(Inkscape) Color is used for the text in Inkscape, but the package 'color.sty' is not loaded}%
    \renewcommand\color[2][]{}%
  }%
  \providecommand\transparent[1]{%
    \errmessage{(Inkscape) Transparency is used (non-zero) for the text in Inkscape, but the package 'transparent.sty' is not loaded}%
    \renewcommand\transparent[1]{}%
  }%
  \providecommand\rotatebox[2]{#2}%
  \newcommand*\fsize{\dimexpr\f@size pt\relax}%
  \newcommand*\lineheight[1]{\fontsize{\fsize}{#1\fsize}\selectfont}%
  \ifx\svgwidth\undefined%
    \setlength{\unitlength}{508.50895381bp}%
    \ifx\svgscale\undefined%
      \relax%
    \else%
      \setlength{\unitlength}{\unitlength * \real{\svgscale}}%
    \fi%
  \else%
    \setlength{\unitlength}{\svgwidth}%
  \fi%
  \global\let\svgwidth\undefined%
  \global\let\svgscale\undefined%
  \makeatother%
  \begin{picture}(1,0.12603829)%
    \lineheight{1}%
    \setlength\tabcolsep{0pt}%
    \put(0,0){\includegraphics[width=\unitlength,page=1]{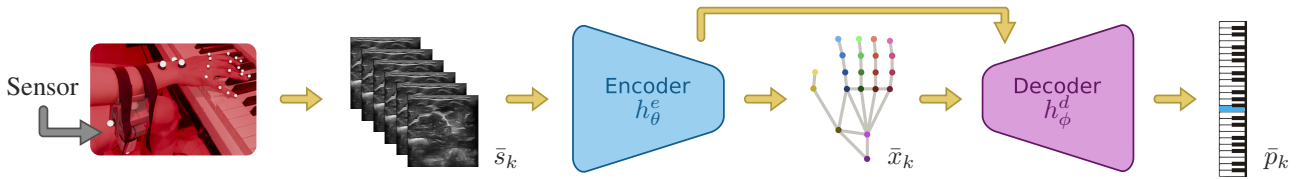}}%
    \put(0.47882401,0.04356086){\color[rgb]{0.06666667,0.36470588,0.54901961}\makebox(0,0)[t]{\lineheight{1.25}\smash{\begin{tabular}[t]{c}$h_{\theta}^e$\end{tabular}}}}%
    \put(0.78234831,0.04356086){\color[rgb]{0.39607843,0.09019608,0.4}\makebox(0,0)[t]{\lineheight{1.25}\smash{\begin{tabular}[t]{c}$h_{\phi}^d$\end{tabular}}}}%
    \put(0.37332394,0.00923141){\color[rgb]{0.15686275,0.15686275,0.15686275}\makebox(0,0)[t]{\lineheight{1.25}\smash{\begin{tabular}[t]{c}$\bar{s}_k$\end{tabular}}}}%
    \put(0.9448878,0.00923141){\color[rgb]{0.15686275,0.15686275,0.15686275}\makebox(0,0)[t]{\lineheight{1.25}\smash{\begin{tabular}[t]{c}$\bar{p}_k$\end{tabular}}}}%
    \put(0.66640355,0.00923141){\color[rgb]{0.15686275,0.15686275,0.15686275}\makebox(0,0)[t]{\lineheight{1.25}\smash{\begin{tabular}[t]{c}$\bar{x}_k$\end{tabular}}}}%
    \put(0.02983458,0.06054252){\color[rgb]{0.15686275,0.15686275,0.15686275}\makebox(0,0)[t]{\lineheight{1.25}\smash{\begin{tabular}[t]{c}Sensor\end{tabular}}}}%
  \end{picture}%
\endgroup%

\vspace{-0.4cm}
\caption{A schematic flow of the proposed model-based method (Sec.~\ref{ssec:indirect-approach}). An ultrasound sensor is placed on the lower arm while the subject is playing the piano. A continuous stream of ultrasound images $\bar{s}_{\ell}$ from the sensor is fed into the neural-network encoder $h^e_{\theta}$ which creates a latent representation of hand skeleton configurations $\bar{x}_{\ell}$. The decoder $h^d_{\phi}$ receives the latent representation and outputs the vector of probabilities $\bar{p}_{\ell}$ indicating the pressed keys. The entire system is trained to produce a common representation for both the skeleton tracking and key prediction tasks.}
\label{fig:diagram}
\vspace{-0.4cm}
\end{figure*}

Specifically, the key question this research aims to answer~is 
\emph{``To what extent can we predict and differentiate between fine finger motions by only having access to the lower-arm muscles?''}
As challenging exemplary use cases, we consider piano playing and keyboard typing.
In these tasks, finger gestures required to press or type are usually finer than those required to grasp objects or fully flex the fingers, generating smaller
lower-arm muscle movements. 
To the best of our knowledge, predicting finger motions for these tasks has not been evaluated as part of an end-to-end system and we hope that our method and dataset will be used as a baseline for understanding these motions.

To this end, we present an end-to-end robotic system that allows to successfully solve our research question. This success stems from the design choices we took to construct our system which we view as a major contribution of this work. These include: (i) choosing an US sensor to capture the muscle's morphological state, (ii) using a sequence of multiple US images at each step, and (iii) a learning framework designed to exploit the temporal nature of the input US images. The proposed neural network (NN) architecture allows to inject domain knowledge accounting for the fact that pressing can be inferred by both the muscles' dynamics and an intermediate lower-dimensional representation of the palm of the hand inspired by robotic manipulators.

As we demonstrate in our empirical evaluation our framework can be used to accurately reproduce music played or text typed, applications that were previously considered unattainable. 
This was done by collecting data from a group of subjects and successfully predicting finger motions in our two motivating applications.\footnote{Note that we do not showcase our results on deformed muscles. We performed tests on healthy subjects with fully-functioning hands in order to provide an accurate benchmark for the task we concentrate on. Our code and dataset are available on \textcolor{blue}{\href{https://github.com/deanzadok/finemotions}{https://github.com/deanzadok/finemotions}}.}
Finally, we demonstrate our system by connecting it to a robotic arm that replayed piano notes played by a subject in real-time.




\section{RELATED WORK}

Object grasping is one of many actions human hands perform daily. One reason for the low adoption rates of prosthetic arms among arm amputees is the lack of functionality in performing daily tasks~\cite{resnik2012advanced}. To enrich the variance of gestures, advanced wearable surface-EMG (sEMG) sensors have been used (e.g., Myo armband~\cite{said2019customizable, DBLP:journals/corr/abs-1801-07756}). The array of features provided by sEMG enabled inferring more than grasp intention, and classifying a varied set of hand gestures via machine learning (ML) and deep learning (DL)~\cite{DBLP:journals/corr/abs-1801-07756, DBLP:conf/hsi/HuangL16}. Additionally, sEMG demonstrated promising performance for arm amputees~\cite{said2019customizable}. Noteworthy, in this domain, subjects differ in their gestures, which puts forth the requirement of operating across a variety of muscular structures~\cite{herbst2020analysis}.

In recent years, US sensing technologies emerged as a promising alternative to EMG. 
%
%
McIntosh et al. proposed to extract the optical flow of an US stream prior to the classification~\cite{DBLP:conf/chi/McIntoshMFP17}. Others used ML algorithms demonstrating gesture classification from downsampled US images of the same area~\cite{DBLP:journals/corr/abs-1808-06543, DBLP:conf/embc/BimbrawFWH20}. Gesture recognition was achieved with wearable single-transducer sensors~\cite{yan2019lightweight, DBLP:journals/tsmc/YangCHYL21}. 
%
%
In addition, researchers demonstrated that similar techniques allow decoupling of different degrees of freedom (DOFs) and use them to classify gestures~\cite{yang2020simultaneous, DBLP:journals/corr/abs-2109-11093}. Unsupervised techniques were also studied, showing discrimination of unlabeled gestures~\cite{DBLP:journals/titb/YangZZHL19}, which might imply the possibility to understand muscle behavior of arm amputees without explicit labeling. Nevertheless, the feasibility of instructing a robotic hand to execute fine finger motions remains an open question.

For the analysis of US imagery DL techniques have become increasingly dominant in tasks ranging from tissue segmentation and pathology detection~\cite{DBLP:journals/access/WangGMQZY21}, and image enhancement~\cite{DBLP:journals/corr/abs-1710-06304}, replacing traditional beamforming techniques with learned beamformers for faster and better image acquisition~\cite{DBLP:journals/corr/abs-1907-02994, DBLP:conf/midl/VedulaSBMZ19}. In the context of learning dynamics of human organs, US imaging with transducers placed under the jaw demonstrated reliable speech and pronunciation recreation from imaged tongue movements~\cite{bliss2018seeing, DBLP:conf/chi/KimuraKR19}. Vogt et al. utilized the tongue's morphological behavior to extrapolate sung pitch and phonemes~\cite{DBLP:conf/nime/VogtMAF02}. Lulich et al. studied the correlation between tongue motion and the frequencies generated by clarinet playing~\cite{lulich2017relation}. These works suggest that US provides a rich signal source describing minute muscular motion.


\section{ANATOMICAL BACKGROUND}
\label{sec:ab}

\begin{figure}[t]
\centering
\subfloat[\label{fig:us_sample_a}]{\includegraphics[trim=0 0 0 0, clip, width=0.138\textwidth]{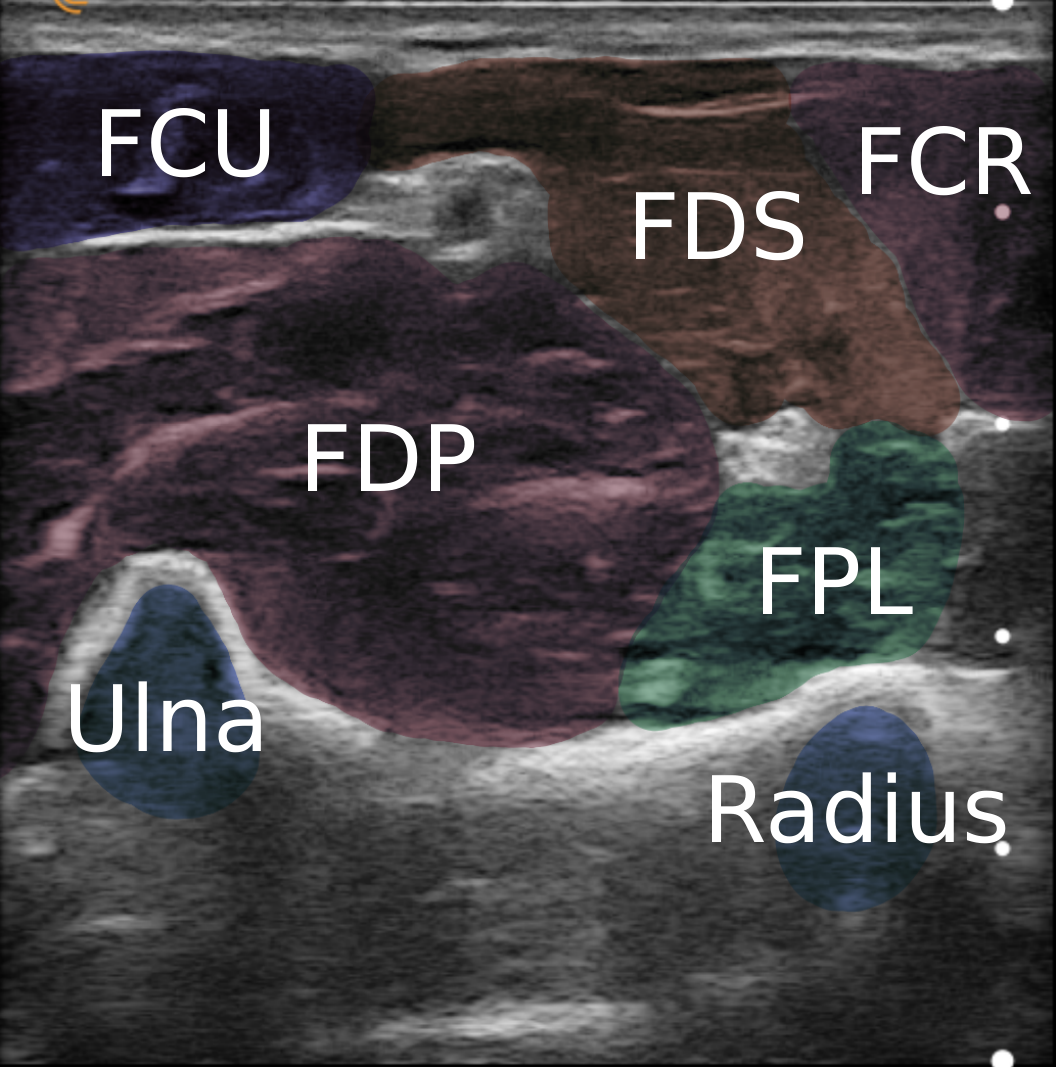}} \hspace{1pt}
\subfloat[\label{fig:us_sample_b}]{\includegraphics[trim=0 0 0 0, clip, width=0.138\textwidth]{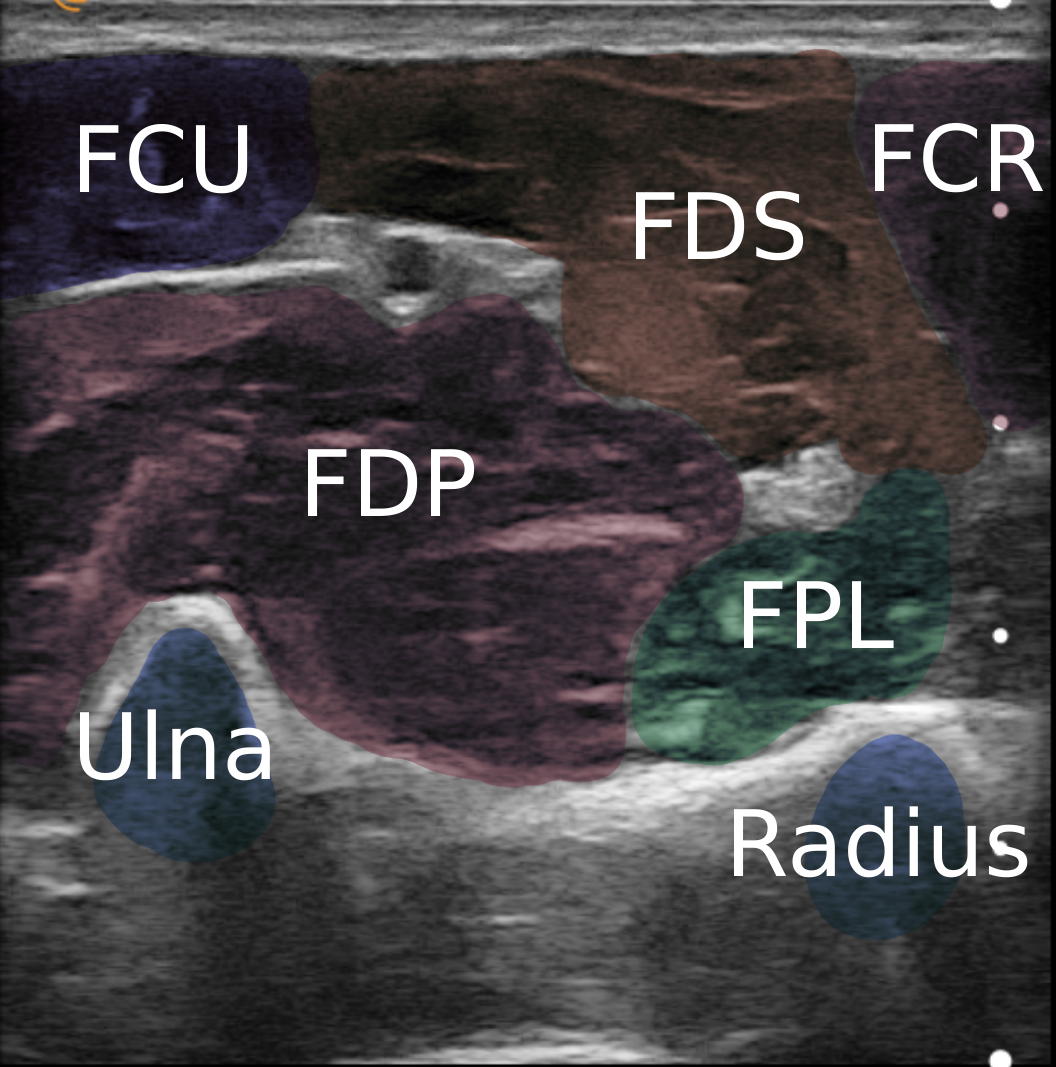}}
\subfloat[\label{fig:conf_def}]{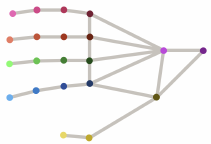}
\vspace{-0.2cm}
\caption{(a,b) Visual segmentation of two US images (top of each image is the skin shell).
The Ulna and Radius (the two main bones), are marked in blue, and the flexion muscles are marked with different colors. 
All fingers are at rest in the left image while in the right image the Ring finger is fully flexed. 
Notice that the region capturing the FDS  in the right image is larger than on the left image as the FDS becomes more dominant in this area during movement. (c)~Visual illustration of a hand's configuration. 
The Thumb~($F^1$) is associated with two joints $[J^1_1,  J^1_2]$,
the other four fingers $F^i$ ($i \in \{2, \ldots, 5\}$) are each associated with three joints $[J^i_1, J^i_2, J^i_3]$ and three additional DOFs are associated with the wrist  $[J^w_r, J^w_p, J^w_y]$. 
The set of joints 
$\{J^i_1 \vert 1 \leq i \leq 5 \}$,
$\{J^i_2 \vert 2 \leq i \leq 5 \}$
and
$\{J^i_3 \vert 2 \leq i \leq 5 \}$
correspond to the MP, PIP and DIP joints, respectively (see Sec.~\ref{sec:ab}).
Figures are best viewed in color.}
\label{fig:us_sample}
\vspace{-0.5cm}
\end{figure}

In this section, we provide the anatomical background required to understand the approach we take to solve our inference problem.
Specifically, we explain (in general terms) the relation between lower-arm muscles and the state of the wrist's and fingers' joints. 
The lower arm contains two main bones, the \emph{Ulna} and the \emph{Radius}.
Surrounding these bones are different muscles that are responsible for the flexion, extension, and rotation of different joints in our palms (including all the fingers). 
Two annotated US images of this region are presented in Fig.~\ref{fig:us_sample_a} and Fig.~\ref{fig:us_sample_b}, highlighting the muscles relevant to our problem as well as different muscle states corresponding to different finger motions. 

We start by concentrating on the four fingers (Index through Little finger).
There are three types of joints associated with these four fingers:
the metacarpophalangeal (MP) joints, 
the proximal interphalangeal (PIP) joints, 
and the distal interphalangeal (DIP) joints (Fig.~\ref{fig:conf_def}).
Close to the skin shell, we can find the \emph{flexor carpi radialis} (FCR) and the \emph{flexor carpi ulnaris} (FCU) that are (together with smaller, less prominent muscles) in charge of flexion of the wrist.
%
Between the FCU and the FCR muscles, we can identify the \emph{flexor digitorum superficialis} (FDS) that flexes the four fingers, and, in full flexion, also contributes to the flexion of the wrist. 
The FDS connects to tendons that go through the PIP and MP joints of the four fingers. 
Deeper into the arm, the \emph{flexor digitorum profundus} (FDP), similar to the FDS, is in charge of finger flexion and wrist-control assistance using tendons connected to the DIP joints. 
Lastly, close to the FDP, is the \emph{flexor pollicis longus} (FPL), which controls the thumb joints.
To summarize, observing the state of the FDS and FDP muscles allows (in theory) to estimate the whereabouts of all four fingers, while the FPL provides the knowledge required to understand thumb movement.

\section{SYSTEM DESIGN}
\label{sec:sd}

Following the anatomical background presented, we argue that one can infer key-pressing actions 
by considering only lower-arm muscles.
To design such a system, we require choosing 
(i)~the type of sensor,
(ii)~the input it provides to the inference algorithm, and
(iii)~the inference algorithm itself.
For our system, we chose to use an US sensor capable of recording 
images at a high frequency, to capture muscle structure.
Choosing an US 
relies on previous work~\cite{DBLP:conf/hsi/HuangL16} in which US completely outperformed EMG in classifying gestures. Moreover, when compared to the single-dimensional signal provided by an EMG sensor, spatial features that are found in an US image have more information about fingers' whereabouts, and are less noisy. This will be key in the way we address the latter two design choices.

Recall that an US provides us with an image. However, in contrast to existing approaches~\cite{DBLP:journals/corr/abs-1808-06543, DBLP:journals/corr/abs-2109-11093}, we chose to use a \emph{sequence} of the latest-acquired US images as the input to our inference algorithm.
As we will demonstrate empirically (Sec.~\ref{sec:exps}), considering a sequence of US images allows to exploit the dynamics which in turn, leads to significantly better results in downstream tasks.

Given a stream of US images we chose to solve the inference problem via a NN architecture that is tailored for processing images, exploits temporal coherence, and allows a domain expert to inject domain knowledge via an intermediate representation.
In our setting, this intermediate representation is an encoding of the hand modeled as a robotic
manipulator.
We detail the exact architecture and specific design considerations in Sec.~\ref{sec:exps}.


\section{INFERENCE LEARNING FRAMEWORK}
\label{sec:methodology}

In this section, we describe our inference learning framework.
We start (Sec.~\ref{ssec:pd}) by formally defining and modeling the inference problem and then continue (Sec.~\ref{ssec:direct-approach}) to describe an approach where we only make use of US images without the aforementioned intermediate representation.
As we will see (Sec.~\ref{sec:exps}), utilizing an intermediate representation during training can lead to better inference capabilities.
Thus, in Sec.~\ref{ssec:indirect-approach} we extend our first method to account for said representation.
Notice that in this section we concentrate on the high-level learning framework and the exact training and inference procedures are detailed in Sec.~\ref{ssec:mat}.

\subsection{Notation and Model}
\label{ssec:pd}

In this section, we introduce the necessary notation and formally define our inference problem.
We assume to be given a sequence of~$k$ US images corresponding to~$k$ timestamps and wish to use them to predict which finger was pressed for each timestamp~$k$.

Let $F^1, \ldots, F^5$ denote the five fingers with~$F^1$ and~$F^5$ corresponding to the Thumb and Little finger, respectively.
For each finger~$F^i$ and each timestamp~$t_j$, we associate a binary variable~$p^i_j \in\{0,1\}$ such that~$p^i_j = 1$ if
finger~$F^i$ is pressed at timestamp~$t_j$ and~$p^i_j = 0$ otherwise.
For every timestamp~$t_j$, we denote by~$p_j \in\{0,1\}^5$ the binary vector indicating which fingers are pressed at~$t_j$ which we term a \emph{pressing vector}.
Namely,~$p_j : = \langle  p^1_j, \ldots, p^5_j \rangle$.
We assume that for every timestamp~$t_j$ there is a corresponding US image~$s_j$ and \emph{configuration}~$x_j$. 
Here, $x_j$ is an encoding of the fingers and wrists as a set of joint angles (see Fig.~\ref{fig:conf_def}).


Our training data consists of tuples~$\langle \bar{s}_\ell, \bar{x}_\ell, \bar{p}_\ell \rangle$.
Here,~$\bar{s}_\ell:= \langle s_{\ell}, \ldots, s_{k+\ell} \rangle$ is a sequence of~$k$ consecutive US images,~$\bar{x}_\ell:= \langle x_{\ell}, \ldots, x_{k+\ell} \rangle$ is a sequence of~$k$ consecutive configurations
and~$\bar{p}_\ell:= \langle p_{\ell}, \ldots, p_{k+\ell} \rangle$ is a sequence of~$k$ consecutive pressing vectors all of which start at timestamp~$t_\ell$.
We denote by~$\calS$ and~$\calX$ the sets of all sequences of~$k$ US images and configurations, respectively.
For each finger~$F^i$ and each timestamp~$t_j$ we wish to predict~$p^i_j$ (i.e., if~$F^i$ was pressed at~$t_j$).
To this end, our system will output the variable $\hat{p}^i_j \in [0,1]$ estimating the probability that~$p^i_j = 1$.
Denote~$\hat{p}_j : = \langle  \hat{p}^1_j, \ldots, \hat{p}^5_j \rangle$ 
and~$\hat{\bar{p}}_\ell:= \langle \hat{p}_{\ell}, \ldots, \hat{p}_{k+\ell} \rangle$.
As we will see, our approach will make use of an intermediate representation corresponding to the configuration at each timestamp.
Thus, we will also denote~$\hat{x}_j$ as the system's estimation of $x_j$ and~$\hat{\bar{x}}_\ell:= \langle \hat{x}_{\ell}, \ldots, \hat{x}_{k+\ell} \rangle$.
In Sec.~\ref{sec:exps} we describe the experimental setting which allows us to obtain labeled training data for our representative tasks.
In the meantime, we describe our approach to solving our inference problem.


\subsection{Learning Pressing Events Directly}
\label{ssec:direct-approach}


To estimate pressing vectors, we use a hybrid architecture starting with a convolutional neural network (CNN) module~\cite{DBLP:journals/corr/GuWKMSSLWW15} to extract spatial features from the US images, followed by a recurrent neural network (RNN)~\cite{rumelhart1985learning, DBLP:journals/corr/abs-1808-03314} to aggregate these features and infer the pressing vectors.
Such a combination was shown to be effective in various computer-vision tasks that require temporal coherence~\cite{guo2021gps, DBLP:journals/complexity/LuLLSW20,DBLP:journals/tvt/ZouJDYCW20}. 
By combining these modules, we create a sequence-to-sequence model, which receives a sequence of US images~$\bar{s}_\ell$ as an input and outputs~$\hat{\bar{p}}_\ell$ which estimate the pressing vector~$\bar{p}_\ell$.
We call this the multi-frame (\MF) model and denote it as~$G_\theta$ where $\theta$ is the set of model parameters that we optimize in the training phase. 
This is done by minimizing the negative log-likelihood, i.e.,
\begin{eqnarray}
\begin{split}
\mathcal{L}_G(\theta) := -\E_{\bar{s}_{\ell} \sim \mathcal{S}}[\log P(\hat{\bar{p}}_{\ell}|\bar{s}_{\ell}, \theta)].
\end{split}
\label{eq:loss_press}
\end{eqnarray}


\subsection{Learning Pressing Events via Hand Configurations}
\label{ssec:indirect-approach}


To estimate pressing vectors while making use of an intermediate representation of the finger's configuration, we use an auto-encoder~\cite{rumelhart1985learning, DBLP:journals/jmlr/Baldi12} framework composed of two main blocks---an encoder mapping the sequence of US images~$\bar{s}_\ell$ to a sequence of configurations~$\hat{\bar{x}}_{\ell}$ and a decoder using~$\hat{\bar{x}}_{\ell}$ to estimate the pressing vector by outputting~$\hat{\bar{p}}_{\ell}$.
The encoder's architecture follows the architecture proposed for the \MF{} model (i.e., a CNN followed by an RNN)
while 
the decoder uses an additional RNN module together with a residual connection~\cite{DBLP:conf/cvpr/HeZRS16} that feeds the model with the hidden features from the encoder
(see Fig.~\ref{fig:diagram} for visualization).

Auto-encoders allow a domain expert to inject domain knowledge via an intermediate representation (hand configurations in our setting)~\cite{DBLP:journals/ral/HuGCLDL21,DBLP:journals/tmm/YangLLL19}.
The residual connection was added to better allow the decoder to infer pressing vectors by obtaining knowledge from both US images and configurations. 
We name this model the Configuration-Based Multi-Frame (\CBMF) model, and denote it as $H_{\theta, \phi} = h^d_{\phi} \circ h^e_{\theta}$  where~$\theta$ and~$\phi$ are the encoder's and decoder's parameters, respectively which we optimize in the training phase.\footnote{Here the superscript `e` and `d` are used to refer to the encoder and decoder modules, respectively.}

While the exact details of how we train this model are explained in the experimental section (Sec.~\ref{ssec:mat}), we note that the training process is done in two phases.
In the first phase,  we train only the encoder~$h^e_{\theta}$ 
while 
in the second phase, we simultaneously train the (pre-trained) encoder~$h^e_{\theta}$ together with the (untrained) decoder~$h^d_{\theta}$.
Splitting the learning into two phases (i.e., a single and a multi-modal supervised learning task) was shown to produce better results in similar tasks~\cite{DBLP:conf/iclr/SinghLZYRL21, tan2019autoencoder, DBLP:conf/icra/ZadokMK21}.
Roughly speaking, by pre-training~$h^e_{\theta}$ we get a ``coarse'' mapping between US images~${\bar{s}_\ell}$ and predicted hand configurations~$\hat{\bar{x}}_\ell$.
This allows us to ``bootstrap'' the network's parameters when solving the entire inference problem in our multi-modal task.
More formally, given $\bar{s}_{\ell}$, $\bar{x}_{\ell}$, and $\bar{p}_{\ell}$, the auto-encoder is trained to predict $\bar{x}_{\ell}$ and $\bar{p}_{\ell}$ given $\bar{s}_{\ell}$ by minimizing the negative log-likelihood of the combination of both terms, i.e.,
\begin{eqnarray}
\begin{split}
\mathcal{L}_H(\theta,\phi)= -\E_{\bar{s}_{\ell} \sim \mathcal{S}, \bar{x}_{\ell} \sim \cal{X}}[&\log P(\hat{\bar{x}}_{\ell}|\bar{s}_{\ell}, \theta) \\
+ &\log P(\hat{\bar{p}}_{\ell}|\hat{\bar{x}}_{\ell}, \bar{s}_{\ell}, \phi)].
\end{split}
\label{eq:loss_multi}
\end{eqnarray}

\section{EXPERIMENTS}
\label{sec:exps}

In this section, we evaluate our approach for predicting fine finger motions.
We start (Sec.~\ref{sec:exps-setup}) by describing our experimental setup and data-collection methodology.
We then describe our model and training details (Sec.~\ref{ssec:mat}) and finish with an analysis of the data collected (Sec.~\ref{ssec:res}).

\subsection{Experimental Setup and Data Collection}
\label{sec:exps-setup}

Our setup includes a Clarius L15HD (wireless B-mode US device), a Roland FP-30 (digital piano), a standard keyboard, and a Vicon motion-capture system. The US's configuration was set for the musculoskeletal task (MSK), with \SI{10}{\mega\hertz} frequency generating a \SI{5}{\cm} $\times$ \SI{5}{\cm} processed image at a rate of $19$-$21$ images per second, in a resolution of $480 \times 480$.
It was attached using a wearable strap (see Fig.~\ref{fig:diagram}) to the so-called ``transverse'' location (see, e.g.,~\cite{DBLP:conf/chi/McIntoshMFP17}). 
The piano and keyboard were used to record the set of pressed notes and keys during each US frame and the Vicon system was used to track the palm's joint locations.
Here, each joint location 
is captured and used to extract the joint angles that form the hand's configuration, as described in \RefAppendix.

We collected data from twelve subjects; averaging at 23 years old, seven of whom are men and five of whom are women. 
All subjects were confirmed to be right-handed, experienced in playing piano, and without any neurological disorders or deformity in the relevant hand area. 
At the beginning of both tasks, subjects were instructed to put their hands in a stationary state in which each of their fingers must be on the same note/key for the entire session. 
For piano playing, subjects were asked to play melodies that only require five distinguished notes and can be played using one hand only. 
For keyboard typing, subjects were asked to randomly type on each of the five keys their fingers are stationed over, while not using more than one key at the same time.
For each task, subjects were asked to perform $2$-$3$ sessions, with approximately ninety seconds for each session, and rest between sessions. The process was approved by the institution's Ethics Committee. The final dataset includes $44,596$ samples for piano playing and $42,143$ samples for keyboard typing, each sample representing the tuple $\langle \bar{s}_\ell, \bar{x}_\ell, \bar{p}_\ell \rangle$ defined in Sec.~\ref{ssec:pd}.

\begin{figure*}[t]
\begin{center}
\includegraphics[trim=0 0 0 0, clip, width=0.85\textwidth]{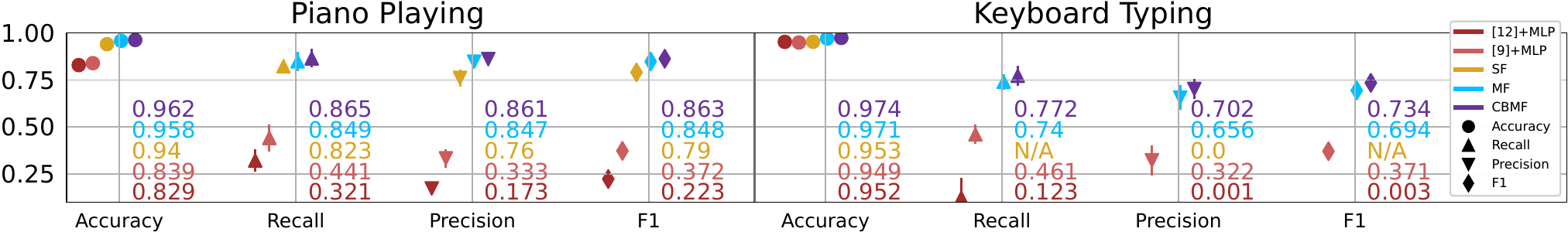}
\end{center}
\vspace{-0.35cm}
\caption{Evaluation of the \MF{} and \CBMF{} methods (Sec.~\ref{ssec:direct-approach} and \ref{ssec:indirect-approach}) as well as the three baselines (Sec.~\ref{ssec:res}) for piano playing and keyboard typing tasks. 
Accuracy, recall, precision, and F1 were averaged over all test samples in $5$ folds. The shape in each interval presents the mean values, while the vertical segment represents the standard deviation range.}
\label{fig:metrics}
\vspace{-0.6cm}
\end{figure*}
\begin{figure*}[t]
\centering
\subfloat[\label{fig:bce_graph_pp}]{\includegraphics[trim=10 8 8 8, clip, width=0.22\textwidth]{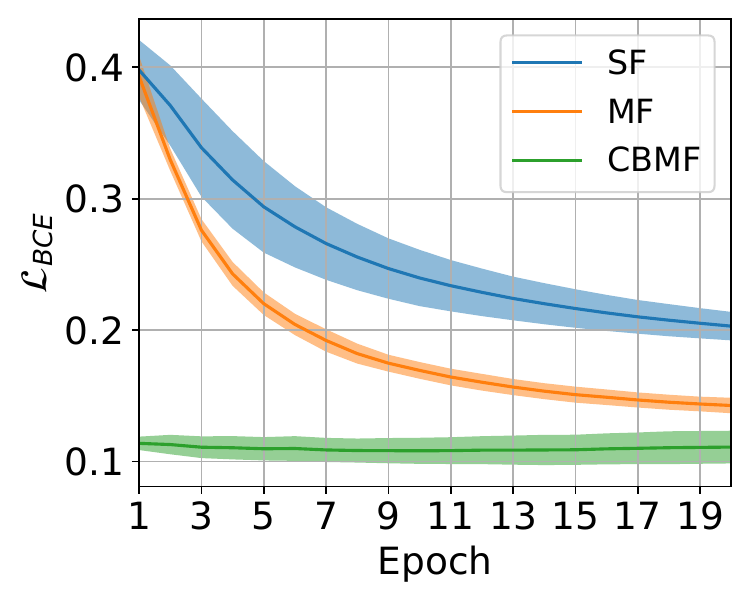}}\hspace{0.1cm}
\subfloat[\label{fig:bce_graph_kt}]{\includegraphics[trim=10 8 8 8, clip, width=0.22\textwidth]{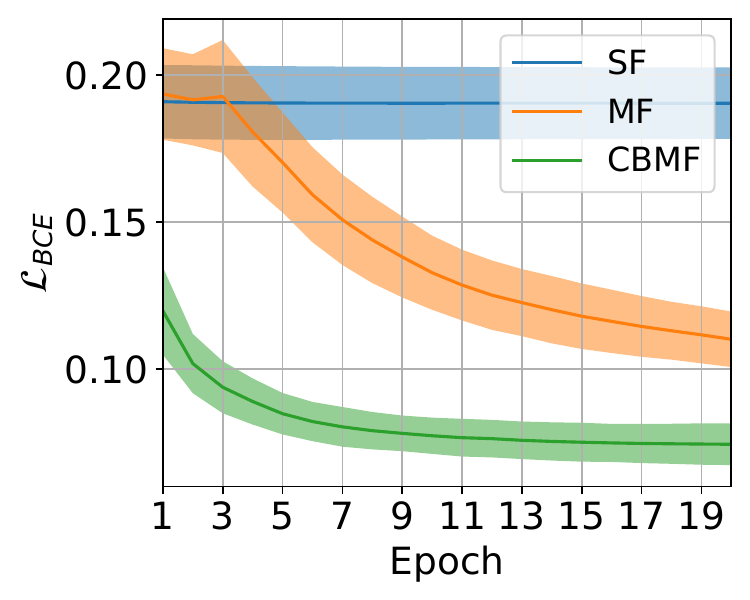}}\hspace{0.1cm}
\subfloat[\label{fig:img_comp_graph}]{\includegraphics[trim=20 10 8 10, clip, width=0.22\textwidth]{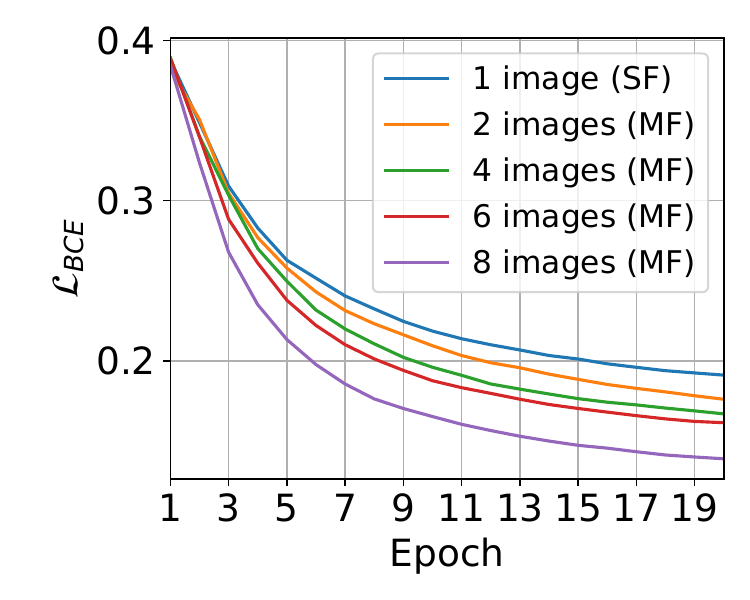}}\hspace{0.15cm}
\subfloat[\label{fig:gen_graph}]{\includegraphics[trim=0 8 8 8, clip, width=0.23\textwidth]{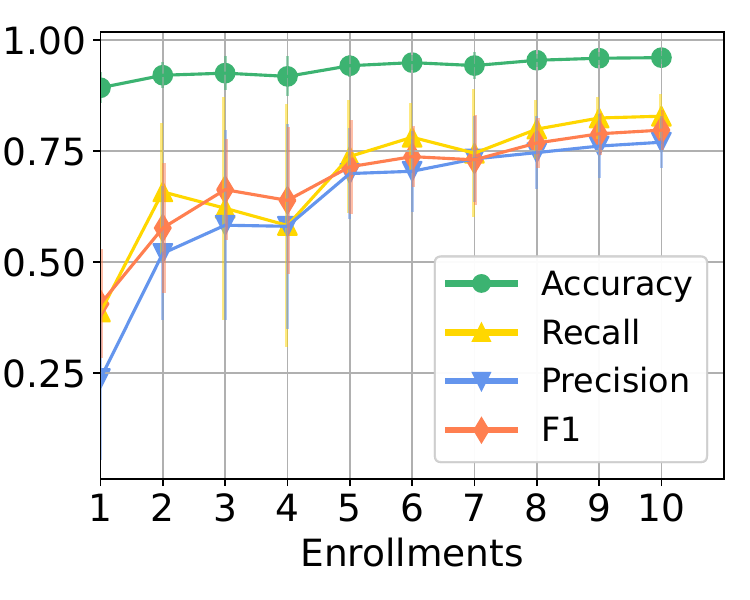}}
\vspace{-0.2cm}
\caption{Comparison of the optimization process using $\mathcal{L}_{\BCE}$ as a function of the epoch number for the piano-playing (a) and keyboard-typing (b) tasks. 
(c)~Training loss as a function of epoch number for different sizes of US sequences as inputs.
(d) Evaluation of the \CBMF{} model on piano playing as a function of the number of enrollments. At each step, we randomly add a new enrollment to the training set and evaluate an unseen enrollment. Results are averaged over $10$ experiments.}
\vspace{-0.5cm}
\end{figure*}

\subsection{Model Training}
\label{ssec:mat}

Before training, both images $s_i$ and configurations $x_i$ were normalized to the range of $[0,1]$. On feedforward, our \MF{} model $G_{\theta}$ receives a sequence of~$k=8$ US images $\bar{s}_\ell$, such that each image~$s_i$ is downsampled to a resolution of $224 \times 224$ and fed to the CNN block to extract feature maps, which are then flattened, concatenated and fed into the RNN module to output~$k$ pressing vectors. Our \CBMF{} model $H_{\theta, \phi}$ works such that the encoder~$h^e_{\theta}$ is being fed with the images similar to the \MF{} model, and outputs $k$ configurations, that are then fed to the decoder~$h^d_{\phi}$ (an additional RNN block), along with the features from the residual connection.
The decoder~$h^d_{\phi}$ is designed to output~$k$ predicted configurations. The entire architecture is detailed in \RefAppendix.
On back propagation, for the predicted configurations, the applied loss function for $N=32$ samples in a batch is the Mean Squared Error (MSE) on all configurations, i.e., $\mathcal{L}_{\MSE} = \frac{1}{N \cdot k} \sum_{i \in N \cdot k} \norm{\hat{x}_i - x_i}_{2}^2.$
A configuration $x_i$ is a vector of $17$ angles, as depicted in Fig.~\ref{fig:conf_def}. For the predicted pressing vectors, the applied loss objective for $N=32$ samples in a batch, such that each sample is $k$ vectors of $M=5$ probabilities, is the Binary Cross Entropy (BCE) for all probabilities, i.e., $\mathcal{L}_{\BCE} = \frac{1}{N \cdot k \cdot M} \sum_{i \in M} \sum_{j \in N \cdot k} p^i_j \log \hat{p}^i_j + (1-p^i_j) \log (1-\hat{p}^i_j)$.
To train the \MF{} model, we apply the loss function $\mathcal{L}_{\BCE}$, and to train the \CBMF{} model, we apply the combined loss function $\lambda \mathcal{L}_{\MSE} + \mathcal{L}_{\BCE}$, where $\lambda$ is chosen using a trial-and-error procedure (see the \RefAppendix). All training sessions were executed for $20$ epochs, to avoid overfitting. Adam~\cite{DBLP:journals/corr/KingmaB14} was used as the optimization method, with a learning rate of $0.001$.

\subsection{Results}
\label{ssec:res}

In all of the experiments, we evaluated all methods using $5$-fold validation: for each task, all sessions of all subjects were divided into five folds, and trained such that the $i$'th fold was set aside for testing and not used in training. This ensures that the test set contains unseen intervals of motions. 

\begin{figure*}[t]
\begin{center}
\setlength\tabcolsep{0pt}
\begin{tabular}{>{\centering\arraybackslash}ccccccccccccccccccc}
\text{ }\text{ }\text{ } & 
\subfloat{\includegraphics[width = 0.058\textwidth]{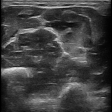}} & 
\subfloat{\includegraphics[width = 0.058\textwidth]{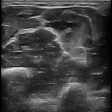}} & 
\subfloat{\includegraphics[width = 0.058\textwidth]{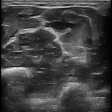}} & 
\subfloat{\includegraphics[width = 0.058\textwidth]{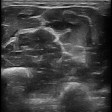}} & 
\subfloat{\includegraphics[width = 0.058\textwidth]{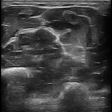}} & 
\subfloat{\includegraphics[width = 0.058\textwidth]{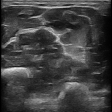}} & 
\subfloat{\includegraphics[width = 0.058\textwidth]{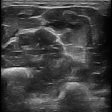}} & 
\subfloat{\includegraphics[width = 0.058\textwidth]{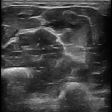}} & \text{ } &\text{ }\text{ }\text{ } & 
\subfloat{\includegraphics[width = 0.058\textwidth]{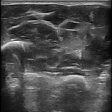}} & 
\subfloat{\includegraphics[width = 0.058\textwidth]{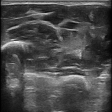}} & 
\subfloat{\includegraphics[width = 0.058\textwidth]{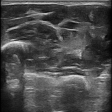}} & 
\subfloat{\includegraphics[width = 0.058\textwidth]{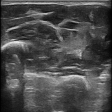}} & 
\subfloat{\includegraphics[width = 0.058\textwidth]{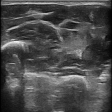}} & 
\subfloat{\includegraphics[width = 0.058\textwidth]{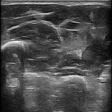}} & 
\subfloat{\includegraphics[width = 0.058\textwidth]{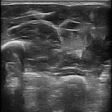}} & 
\subfloat{\includegraphics[width = 0.058\textwidth]{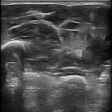}} \\ [-3.0ex]
\multicolumn{9}{l}{\subfloat{\includegraphics[trim=0 21 5 0, clip, width=0.483\textwidth]{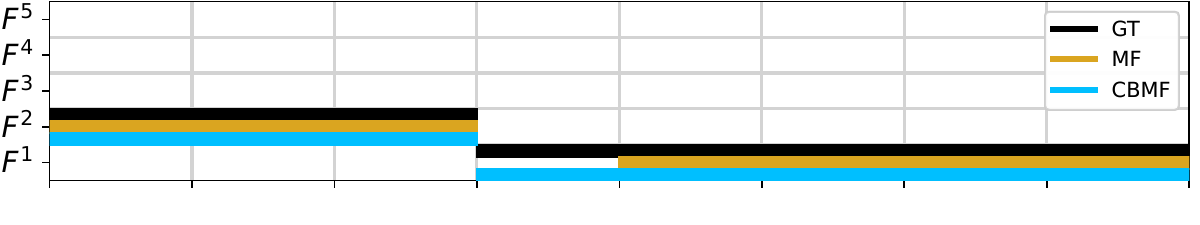}}} & \text{ } & \multicolumn{9}{l}{\subfloat{\includegraphics[trim=0 21 5 0, clip, width=0.483\textwidth]{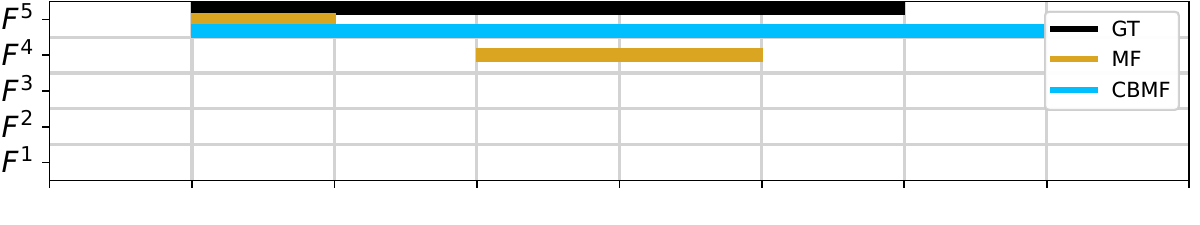}}} \\ [-2.8ex]
\end{tabular}
\end{center}
\caption{Samples from executing the \MF{} and \CBMF{} models. In each rollout, the upper row represents finger $F^5$ (Little finger) and the lower row represents finger $F^1$ (Thumb). Ground-truth (GT) labels are marked in black. The differences between both models are expressed in consistency and the number of false-positive predictions. Notice from the US images how hard is it for a human eye to understand subtle morphological changes during such gentle movements.}
\label{fig:samples}
\vspace{-0.6cm}
\end{figure*}

\begin{figure*}[t]
\centering
\subfloat{\includegraphics[trim=0 0 0 0, clip, width=0.63\textwidth]{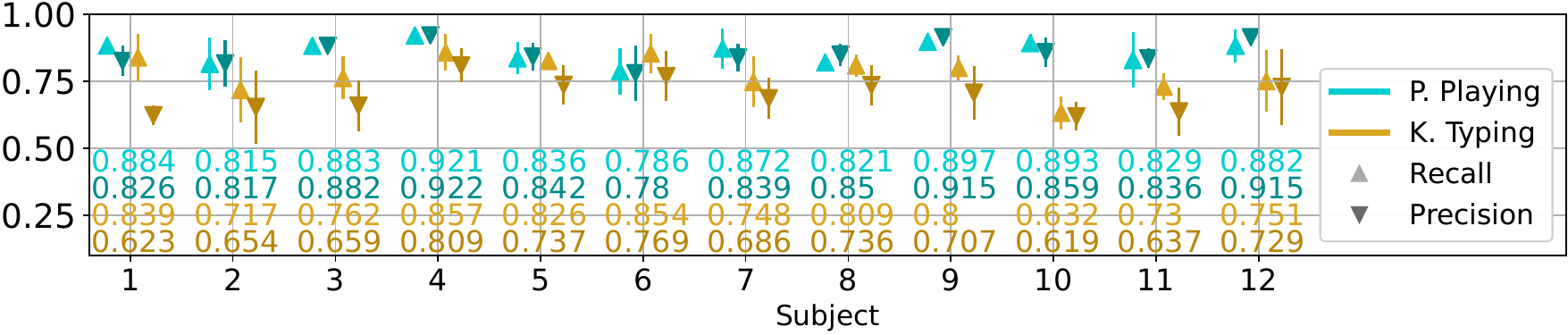}}\hspace{0.2cm}
\subfloat{\includegraphics[trim=0 0 0 0, clip, width=0.338\textwidth]{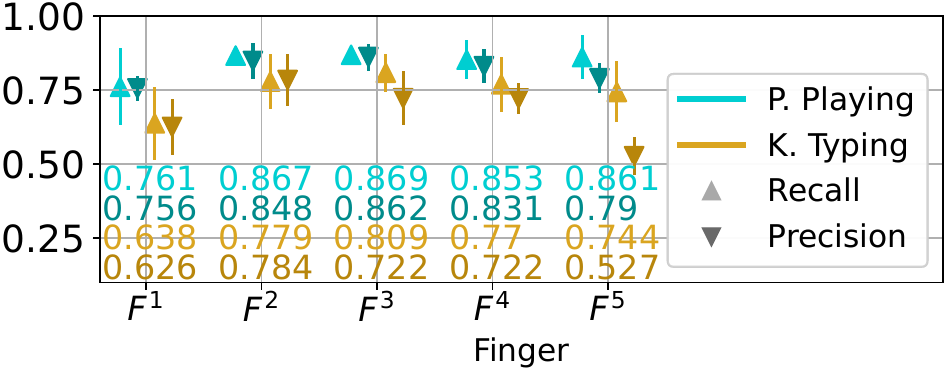}}
\vspace{-0.2cm}
\caption{(Left) Recall and precision for the \CBMF{} model for all twelve subjects. Vertical segments denote one standard deviation. 
(Right) Recall and precision separated for all five fingers, showing the expected performance of each finger.
}
\label{fig:metrics_subjectsfingers}
\vspace{-0.25cm}
\end{figure*}

\begin{figure*}[t]
\centering
\includegraphics[trim=0 6 0 0, clip, width = 0.90\textwidth]{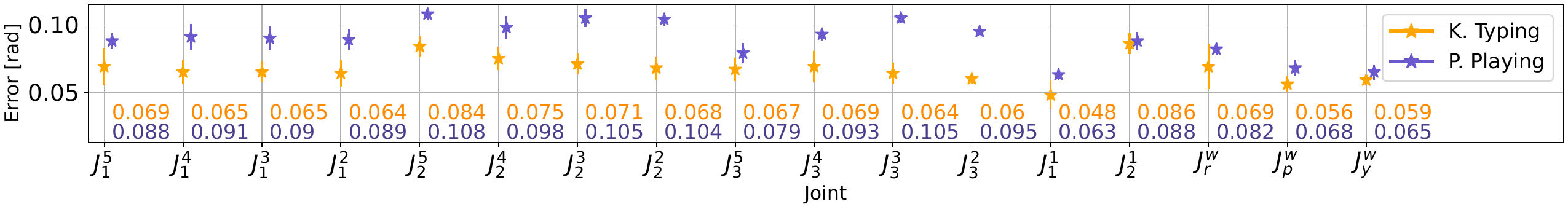}
\vspace{-0.15cm}
\caption{Error-values in radians for predicting joint angles using our \CBMF{} model.}
\label{fig:joints}
\vspace{-0.5cm}
\end{figure*}

\textbf{How good are the proposed methods?} 
We evaluated the inference capabilities of our two models (\MF{} and \CBMF{}) on the piano-playing and keyboard-typing tasks by plotting the accuracy, recall, precision, and $F1$ (Fig.~\ref{fig:metrics}) as well as the test loss during training (Fig.~\ref{fig:bce_graph_pp} and \ref{fig:bce_graph_kt}).
For baselines, we added the single-frame (\SF{}) model, where we simply take our \MF{} model and replace the RNN with a multilayer perceptron (MLP) to predict a single pressing vector from a single image. 
In addition, we implemented state-of-the-art algorithms for gesture classification~\cite{DBLP:conf/chi/McIntoshMFP17,DBLP:conf/hsi/HuangL16}. 
To allow the detection of pressing vectors, we concatenated an MLP to their output.
Implementation details are in \RefAppendix.
We also tested common computer-vision models, such as the VGG16~\cite{DBLP:journals/corr/SimonyanZ14a, DBLP:journals/corr/abs-2109-11093} model (pretrained on ImageNet~\cite{DBLP:conf/cvpr/DengDSLL009}, or not), and C3D~\cite{DBLP:conf/iccv/TranBFTP15}, which were shown to perform favorably in video classification. However, the models did not converge in any of the tasks and therefore were excluded from evaluation.


The overall performance for piano playing is significantly higher than for keyboard typing.
We associate this difference with differences in pressing duration and body postures between the tasks:
The maximum note offset on our piano and our keyboard is \SI{10.45}{\mm} and \SI{2.95}{\mm}, respectively. 
This implies longer muscle flexing when playing the piano, which leads to more dominant motions in US images. 
Additionally, playing the piano involves different body postures which often lead to more dominant motions in US images. 

Moving on to comparing the different models, the \SF{} model is unable to recreate the finer task of keyboard typing, but is able to learn to play the piano.
This backs up our conjecture that a sequence of multiple frames is required to infer fine finger motions.
Notice that the \MF{} model outperforms the three baselines while the \CBMF{} model outperforms all.
This is especially noticeable in the harder keyboard-typing task (see recall and precision), which backs up our conjecture that our intermediate representation allows us to better infer fine finger motions, by reducing false positive and false negative predictions.
%
%
In addition, we examined piano playing with sequences of images of different sizes ($k \in \{1, 2, 4, 6, 8\}$), showing that convergence on the test set improves gradually the more we increase $k$.
Larger sequences may have been beneficial, however, we did not use more than $8$ consecutive images due to GPU memory limitations.
%
For a representative visualization of the difference between the \MF{} and \CBMF{} models, see Fig.~\ref{fig:samples}.

\textbf{How feasible is the method for a group of subjects?} 
We plot the different metrics for each subject individually to see how the \CBMF{} model varies across different subjects (Fig.~\ref{fig:metrics_subjectsfingers}). 
Overall, the results for piano playing are better than for keyboard typing, and those who performed better on piano playing, performed better on keyboard typing, but most importantly, the method works for all subjects, showing us the potential for large-scale adoption of these algorithms.


\textbf{How well does the system works for all fingers?} We computed the recall and precision values for each of the fingers individually to understand how each of the fingers affects the quality of the solution (Fig.~\ref{fig:metrics_subjectsfingers}). 
On both tasks, the model performs better on fingers $F^2, F^3$ and $F^4$, rather than on fingers $F^1$ and $F^5$. 
A possible explanation is that operating the three middle fingers causes larger movement in the muscles that are found in the US images when compared to the Thumb and Little finger. 
As illustrated in Fig.~\ref{fig:us_sample}, the muscle responsible for Thumb movement  is significantly less dominant than the rest of the muscles, which might explain why Thumb movement is the least-understandable gesture.

\textbf{How well can the model generalize on real-time data?} 
Our method works without any preprocessing, and can work in an online setting.
This is in contrast to existing methods for which subjects were instructed to perform gestures according to predefined instructions~\cite{DBLP:journals/corr/abs-1808-06543}, or separate between gesture categories before feature extraction~\cite{DBLP:conf/chi/McIntoshMFP17}.

\begin{figure}[h]
\centering
\includegraphics[trim=0 0 0 0, clip, width = 0.40\textwidth]{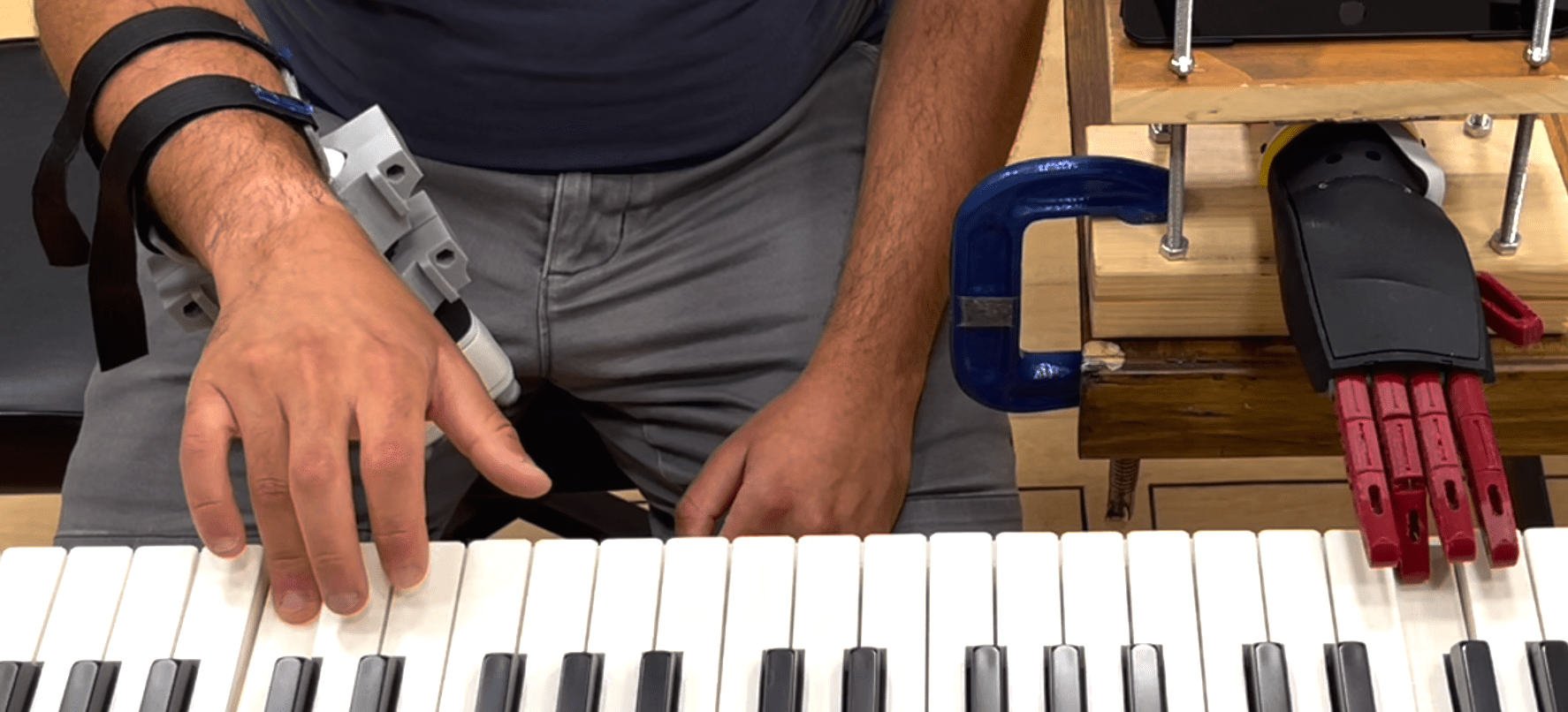}
\vspace{-0.15cm}
\caption{Our system connected to a robotic hand.}
\label{fig:robotic_hand}
\vspace{-0.5cm}
\end{figure}

A potential challenge in real-time inference is that the US sensor is not placed in the exact location for which training data was obtained (also known as sensor reattachment). To this end, we define ``enrollment'' as a period of $2$-$3$ recording sessions, in which the US sensor was not removed from the hand.
We used a single subject to demonstrate the number of enrollments needed to infer from an unseen enrollment. 
Results (Fig.~\ref{fig:gen_graph}) show that 
(i) as the number of enrolments increases in the training session, our inference capabilities increase 
and
(ii) the number of enrolments required in the training session is relatively moderate.
We connected our system to a robotic hand (see Fig.~\ref{fig:robotic_hand} as well as the supplementary video) which continuously replayed in real time notes played on the piano.

\textbf{How accurate are the estimated configurations?} 
We evaluated the error values (in radians) of our \CBMF{} model for each of the individual $17$ joints in a configuration (see Fig.~\ref{fig:conf_def}).
Results, depicted in Fig.~\ref{fig:joints}, demonstrate that the more the task requires longer movements, the higher the average error, especially for the more dominant joints, such as fingers~$F^2$ and $F^3$. 
Interestingly, the DOFs that are more dominant near the wrist ($J^1_1, J^w_r, J^w_p, J^w_y$), achieve a smaller error on average, which might relate to the large region captured by the relative muscles in the US images.



%
%



\section{DISCUSSION AND FUTURE WORK}
\label{sec:dac}

In this paper, we explored the task of predicting and differentiating between fine finger motions by only having access to a stream of lower-arm US images.
We show that incorporating temporal coherence and a kinematic hand representation in data-driven methods is essential for solving such tasks.
We believe that the inference capabilities we present will pave the way towards enabling fully-operational prostheses for those in need.
Importantly, generalization for unseen subjects was not dealt with in this paper, as it requires holding prior knowledge regarding the muscles' structure.

Future directions that we are exploring include
(i) Restoring hand motions for people with muscular deformations such as amputees. This will potentially require data-driven solutions to learn continuous and unlabeled finger motions for non-able-bodied subjects.
(ii) Recent advancements in feedback sensing
allow amputees to sense objects they are engaging with~\cite{hari2021advanced}. Sensor feedback can also be used to endow our inference algorithm with additional input such as surface type 
which can be used to improve decision making.
%

\ifthenelse{\boolean{isARXIVversion}}{

\section*{ACKNOWLEDGMENT}
\label{sec:ack}


This project has received funding from 
the European Research Council (ERC) under the European Union’s Horizon 2020 research and innovation programme (grant agreement No. 863839),
the Israeli Ministry of Science \& Technology grants No. 3-16079 and 3-17385,
and the United States-Israel Binational Science Foundation (BSF) grants no. 2019703 and 2021643.
We also thank Haifa3D, a non-profit organization providing self-made 3d-printed solutions, for their consulting and support through the research.

}{}

\ifthenelse{\boolean{isARXIVversion}}{

\section*{APPENDIX}
\label{sec:app}


\subsection*{Joints Computation}
\label{ssec:jc}


The set of joint angles forming a configuration is acquired by placing markers on the relevant joints and using a motion-capture system to obtain the $[x, y, z]$ location  of each marker {(see Fig.~\ref{fig:markers_sample})}. 
Then, each  joint angle is computed using  the markers' locations as we detail next. 
Hereafter, we compute vectors as the point-wise subtraction of the marker locations.
For example, given two 3D points (that can be considered as two 3D vectors w.r.t. some reference frame) $f_{42}$ and~$f_{43}$, we denote the vector from $f_{42}$ to $f_{43}$ as follows:

\begin{eqnarray}
\begin{split}
\Vec{v}_{f_{42} \rightarrow f_{43}} &= \Vec{f_{43}} - \Vec{f_{42}}.
\end{split}
\label{eq:vector_example}
\end{eqnarray}

Next, we describe how we compute the angle between two vectors. 
Consider, for example,  the vector from $f_{42}$ to~$f_{43}$ and the vector from $f_{42}$ to~$f_{41}$, we compute the joint angle~$J^5_2$ as follows:

\begin{eqnarray}
\begin{split}
J^5_2 = \arctantwo ( \norm{&\Vec{v}_{f_{42} \rightarrow f_{41}} \times \Vec{v}_{f_{42} \rightarrow f_{43}}}_2 ,\\
&\Vec{v}_{f_{42} \rightarrow f_{41}} \cdot \Vec{v}_{f_{42} \rightarrow f_{43}}).
\end{split}
\label{eq:angle_example}
\end{eqnarray}

This allows us to compute joint angles. 
We start with  the four fingers $F^2, \ldots, F^5$ and will use finger~$F^5$ (Little finger) for example. 
The three joints that are required   are $J^5_1$, $J^5_2$ and $J^5_3$. 
Hence, we will compute $J^5_1$ and $J^5_3$ as follows (fingers~$F^1$, $F^2$, and~$F^3$ are calculated in a similar fashion):

\begin{eqnarray}
\begin{split}
J^5_1 = \arctantwo ( \norm{&\Vec{v}_{f_{41} \rightarrow t_{5}} \times \Vec{v}_{f_{41} \rightarrow f_{42}}}_2 , \\
&\Vec{v}_{f_{41} \rightarrow t_{5}} \cdot \Vec{v}_{f_{41} \rightarrow f_{42}}), \\
J^5_3 = \arctantwo ( \norm{&\Vec{v}_{f_{42} \rightarrow f_{41}} \times \Vec{v}_{f_{42} \rightarrow f_{43}}}_2 , \\
&\Vec{v}_{f_{42} \rightarrow f_{41}} \cdot \Vec{v}_{f_{42} \rightarrow f_{43}}).
\end{split}
\label{eq:angles_4143}
\end{eqnarray}

As for the Thumb, we compute two joint angles; $J^1_1$ and~$J^1_2$. 
Similar to Eq.~\ref{eq:angles_4143}, we compute $J_2^1$ as follows:

\begin{eqnarray}
\begin{split}
J^1_2 = \arctantwo ( \norm{&\Vec{v}_{t_{6} \rightarrow t_{7}} \times \Vec{v}_{t_{6} \rightarrow t_{5}}}_2 , \\
&\Vec{v}_{t_{6} \rightarrow t_{7}} \cdot \Vec{v}_{t_{6} \rightarrow t_{5}}),
\end{split}
\label{eq:angles_t6}
\end{eqnarray}

\noindent where joint $J^1_1$ is acquired by computing the angle between~$\Vec{v}_{t_{5} \rightarrow t_{6}}$ and the normal vector to the plane approximating the palm's plane, by taking the three points $f_{11}$, $f_{41}$ and $t_4$. Denoting this normal as $\hat{\Vec{n}}_p$, joint $J^1_1$ is computed as follows:

\begin{eqnarray}
\begin{split}
J^1_1 = \arctantwo ( \norm{&\Vec{v}_{t_{5} \rightarrow t_{6}} \times \hat{\Vec{n}}_p}_2 , \\
&\Vec{v}_{t_{5} \rightarrow t_{6}} \cdot \hat{\Vec{n}}_p).
\end{split}
\label{eq:angles_t5}
\end{eqnarray}

This is done differently in order to compute the angle that corresponds to the pressing motion of the Thumb. 
As for the three DOFs of the wrist, we have $J^w_r$, $J^w_p$, and $J^w_y$ for roll, pitch and yaw respectively. As for pitch, we compute the angle between $\hat{\Vec{n}}_p$ and $\Vec{v}_{t_{4} \rightarrow t_{3}}$. For yaw, we compute the angle between $\Vec{v}_{f_{41} \rightarrow f_{11}}$ and $\Vec{v}_{t_{4} \rightarrow t_{3}}$, and for roll, we compute the angle between $\hat{\Vec{n}}_p$ and a vector obtained from two additional points located on the elbow.

\begin{figure}[t]
\centering
\includegraphics[trim=0 10 0 20, clip, width = 0.30\textwidth]{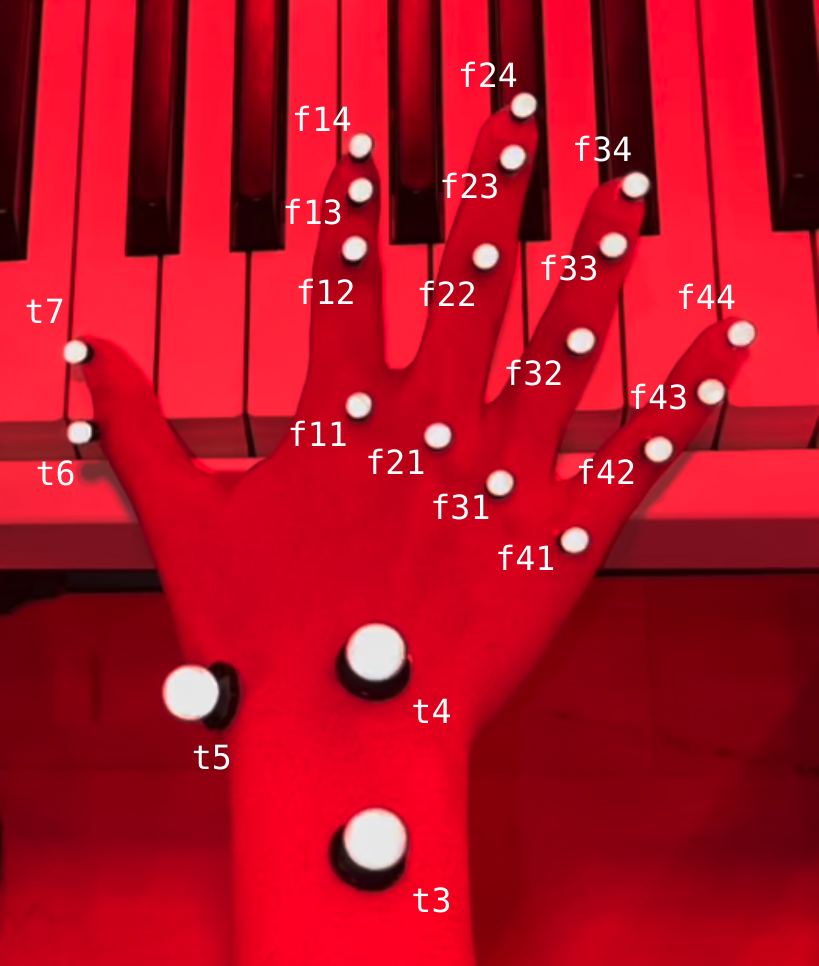}
\caption{An image showing the position of each marker on the subject's hand.}
\label{fig:markers_sample}
\vspace{-0.3cm}
\end{figure}


\subsection*{Network Architecture}
\label{ssec:na}


Tables~\ref{tab:mf_arch} and Table~\ref{tab:cbmf_arch} shows the network architectures for the \MF{} and \CBMF{} models, respectively. The tables are divided into blocks, such that the first block represents the CNN module and the others represent the RNN modules. The input shape for both models is $[224 \times 224 \times 8]$, and corresponding to $8$ consecutive grayscale US images. The shape of the intermediate representation is $[8 \times 17]$, corresponding to a trajectory of $8$ robotic configurations, each of size $17$. The output shape is $[8 \times 5]$, denoting $8$ pressing vectors, each containing five probabilities for the five fingers.

The CNN module is based on the encoding part of UNet~\cite{DBLP:conf/miccai/RonnebergerFB15}, due to the high success of this architecture when used in medical imaging tasks. This block is used to extract $512$ $3 \times 3$ feature maps from each of the eight images, such that the same parameters are shared across all images. These are flattened, concatenated, and fed into the recurrent block. In the \CBMF{} model, The trajectory, along with $8 \times 128$ features are taken from the output of the second GRU layer, concatenated, and fed as an input to the decoder.

\begin{table}[h]
\centering
\begin{tabular}{lccc}  
\toprule
Layer & Act. & Out. shape & Parameters \\
\midrule
Conv2D & ReLU & $224 \times 224 \times 32$ & 320 \\ 
Conv2D & ReLU & $224 \times 224 \times 32$ & 9,248 \\ 
MaxPooling2D ($2 \times 2$) & - & $112 \times 112 \times 32$ & - \\ 
Conv2D & ReLU & $112 \times 112 \times 64$ & 18.50k \\ 
Conv2D & ReLU & $112 \times 112 \times 64$ & 36.93k \\ 
MaxPooling2D ($2 \times 2$) & - & $56 \times 56 \times 64$ & - \\ 
Conv2D & ReLU & $56 \times 56 \times 128$ & 73.86k \\ 
Conv2D & ReLU & $56 \times 56 \times 128$ & 147.58k \\ 
MaxPooling2D ($2 \times 2$) & - & $28 \times 28 \times 128$ & - \\ 
Conv2D & ReLU & $28 \times 28 \times 256$ & 295.17k \\ 
Conv2D & ReLU & $28 \times 28 \times 256$ & 590.08k \\ 
Dropout (p=0.3) & - & $28 \times 28 \times 256$ & - \\ 
MaxPooling2D ($2 \times 2$) & - & $14 \times 14 \times 256$ & - \\ 
Conv2D & ReLU & $14 \times 14 \times 512$ & 1.18M \\ 
Conv2D & ReLU & $14 \times 14 \times 512$ & 2.36M \\ 
Dropout (p=0.3) & - & $14 \times 14 \times 512$ & - \\ 
MaxPooling2D ($4 \times 4$) & - & $3 \times 3 \times 512$ & - \\ 
Flatten  & - & $4608$ & - \\
\midrule
Concatenate ($8$ images) & - & $8 \times 4608$ & - \\
GRU & ReLU & $8 \times 1024$ & 17.31M \\ 
GRU & ReLU & $8 \times 128$ & 443.14K \\ 
GRU & Linear & $8 \times 5$ & 2,025 \\
\midrule
Total & & & 22.93M \\
\toprule 
\end{tabular}
\caption{Neural-network architecture of our \MF{} model~$G_{\theta}$.}
\label{tab:mf_arch}
\end{table}

\begin{table}[h]
\centering
\begin{tabular}{lccc}  
\toprule
Layer & Act. & Out. shape & Parameters \\
\midrule
Conv2D & ReLU & $224 \times 224 \times 32$ & 320 \\ 
Conv2D & ReLU & $224 \times 224 \times 32$ & 9,248 \\ 
MaxPooling2D ($2 \times 2$) & - & $112 \times 112 \times 32$ & - \\ 
Conv2D & ReLU & $112 \times 112 \times 64$ & 18.50k \\ 
Conv2D & ReLU & $112 \times 112 \times 64$ & 36.93k \\ 
MaxPooling2D ($2 \times 2$) & - & $56 \times 56 \times 64$ & - \\ 
Conv2D & ReLU & $56 \times 56 \times 128$ & 73.86k \\ 
Conv2D & ReLU & $56 \times 56 \times 128$ & 147.58k \\ 
MaxPooling2D ($2 \times 2$) & - & $28 \times 28 \times 128$ & - \\ 
Conv2D & ReLU & $28 \times 28 \times 256$ & 295.17k \\ 
Conv2D & ReLU & $28 \times 28 \times 256$ & 590.08k \\ 
Dropout (p=0.3) & - & $28 \times 28 \times 256$ & - \\ 
MaxPooling2D ($2 \times 2$) & - & $14 \times 14 \times 256$ & - \\ 
Conv2D & ReLU & $14 \times 14 \times 512$ & 1.18M \\ 
Conv2D & ReLU & $14 \times 14 \times 512$ & 2.36M \\ 
Dropout (p=0.3) & - & $14 \times 14 \times 512$ & - \\ 
MaxPooling2D ($4 \times 4$) & - & $3 \times 3 \times 512$ & - \\ 
Flatten  & - & $4608$ & - \\
\midrule
Concatenate ($8$ images) & - & $8 \times 4608$ & - \\
GRU & ReLU & $8 \times 1024$ & 17.31M \\ 
GRU & ReLU & $8 \times 128$ & 443.14K \\ 
GRU & Linear & $8 \times 17$ & 7,497 \\
\midrule
Merge & - & $8 \times 145$ & - \\
GRU & ReLU & $8 \times 256$ & 309.50K \\ 
GRU & ReLU & $8 \times 128$ & 148.22K \\ 
GRU & Linear & $8 \times 5$ & 2,025 \\ 
\midrule
Total & & & 22.93M \\
\toprule 
\end{tabular}
\caption{Neural-network architecture of our \CBMF{} model~$H_{\theta,\phi}$.}
\label{tab:cbmf_arch}
\end{table}

\begin{figure}[h]
\centering
\includegraphics[width = 0.37\textwidth]{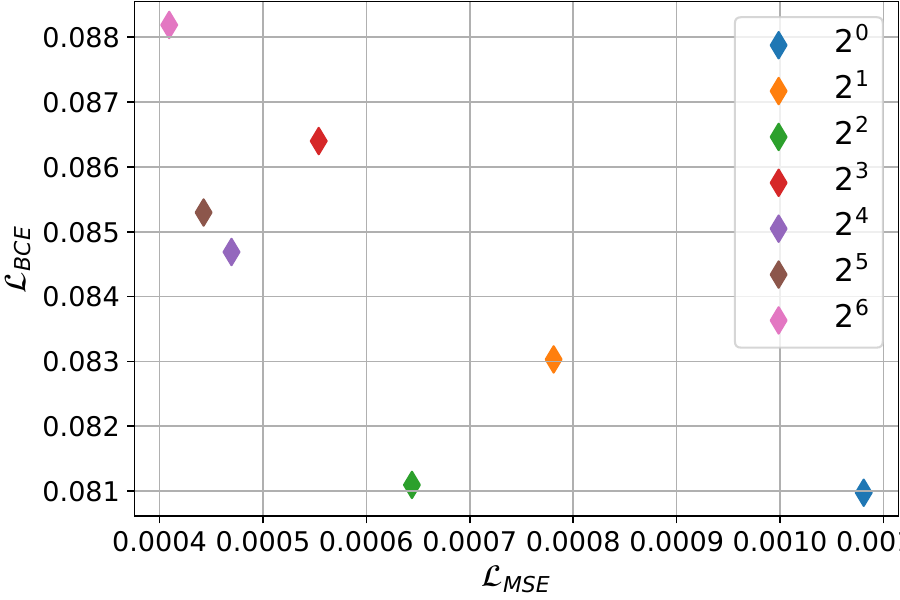}
\caption{$\mathcal{L}_{\BCE}$ and $\mathcal{L}_{\MSE}$ for different values of $\lambda$. On average, results are showing that using lower values of $\lambda$ yields lower values of $\mathcal{L}_{\BCE}$ and higher values of $\lambda$ yields lower values of $\mathcal{L}_{\MSE}$ and higher values of $\mathcal{L}_{\BCE}$.}
\label{fig:losses_weighting}
\end{figure}


\subsection*{Baselines Implementation}
\label{ssec:baseimp}


As presented in Sec.~\ref{ssec:res}, we implemented three baselines to asses our models in solving the two downstream tasks. The evaluation was using the same $5$-fold validation over the same datasets. 
We started by evaluating the importance of temporal coherence by defining the \SF{} model. Here, we take our \MF{} model and replace the RNN with a MLP, such that the CNN module is the same as in the \MF{} and \CBMF{} models, and the addition is two linear layers. On inference, the model receives a single $224 \times 224$ US image and predict a single pressing vector.

In addition we compared our work to two additional papers that succeeded in gesture classification.
Importantly, these papers were originally designed to classify full recorded sessions in an offline setting, and since these algorithms were evaluated on different gestures with different sensors, we modified the input to receive the last acquired $8$ images as in our models.
Second, these algorithms were designed to discriminate between gestures and here we allow multiple fingers pressing at the same time. Therefore, we set the amount of available classes to $2^5=32$ (i.e., a class corresponds to a subset of fingers pressed).

Huang et al.~\cite{DBLP:conf/hsi/HuangL16} proposed to sample five columns of pixels from the image, imitating the feature structure of A-mode transducers. These columns are then separated into segments, from which they extracted two coefficients using linear fitting. Then, these features are concatenated across the temporal axis and fed to the classifier. Here, the authors used sequences of $36$ images, but since we did not observe any improvement, the results are presented for sequences of eight images. 
Next, McIntosh et al.~\cite{DBLP:conf/chi/McIntoshMFP17} proposed to extract the optical flow from adjacent frames, perform average pooling on patches of size $20 \times 20$, and accumulate them across the temporal axis. In the same way, we calculated these features for a sequence of eight US images and fed them to the classifier. 
For both baselines, we evaluated all classifiers that were proposed by the authors, and in both cases, MLP outperformed all other classifiers.



\subsection*{Losses Weighting}
\label{ssec:lw}


We empirically chose the value of $\lambda$ to minimize the objective function $\lambda \mathcal{L}_{\MSE} + \mathcal{L}_{\BCE}$. Here, we executed training sessions to obtain results for $\lambda \in \{ 2^i | i \in [0, \hdots, 6] \}$ and chose the result yielding the lowest $\mathcal{L}_{\BCE}$, without harming the optimization of $\mathcal{L}_{\MSE}$. Results, presented in Fig.~\ref{fig:losses_weighting}, show that by taking $\lambda = 4$, we benefit from using both modalities. The same experiment with $\lambda = 0$ (i.e., only using $\mathcal{L}_{\BCE}$) resulted in $\mathcal{L}_{\BCE} = 0.093$ and $\mathcal{L}_{\MSE} = 2.582$, and therefore was ruled out from the evaluation.


\subsection*{Real-Time Implementation}
\label{ssec:rti}


We built a   setup for online real-time inference to demonstrate piano playing on demand. The code is written in \Cpp{} and includes multi-threading to deal with interfaces to the US sensor, the robotic hand, and model inference simultaneously. For optimized inference, our \CBMF{} model was divided into two software modules - the encoder's convolution block and the remaining recurrent blocks including the residual connection. On inference, the incoming images from the US sensor are fed to the first module to extract features, while an additional thread, simultaneously feeds the sequence of the last eight extracted features to the second module, and stores the predictions as instructions that are sent to the robotic hand.
The model was optimized using the ONNX Runtime library, the entire model weighs \SI{745}{MB} and the inference is executed at \SI{60}{\hertz} on a powerful GPU (Nvidia RTX 2080Ti) or \SI{29}{\hertz} on a low-budget GPU (Nvidia Quadro P620).
}{}




{\small
\bibliographystyle{IEEEtran}
\bibliography{main}
}

\end{document}